\documentclass[runningheads]{llncs}

\usepackage[square,sort,comma,numbers]{natbib} 
\usepackage{xcolor}
\usepackage{graphicx}
\usepackage{caption}
\usepackage{subfig}
\usepackage{stackengine}
\usepackage{amssymb}
\usepackage{amsmath}
\usepackage{mathtools}
\usepackage{bbm}
\usepackage{booktabs}
\usepackage{multirow}
\usepackage{makecell}
\begin{document}
%
\title{Unsupervised Cross-domain Image Classification by Distance Metric Guided Feature Alignment}
%
\titlerunning{Unsupervised Cross-domain Image Classification}
%
\author{
  Qingjie Meng\inst{1}
  \and Daniel Rueckert\inst{1}
  \and Bernhard Kainz\inst{1}
}
%
\authorrunning{Q. Meng et al.}
%
\institute{Department of Computing, BioMedIA, Imperial College London, UK\\
\email{q.meng16@imperial.ac.uk}
}
%
\maketitle              
\begin{abstract}
Learning deep neural networks that are generalizable across different domains remains a challenge due to the problem of domain shift. Unsupervised domain adaptation is a promising avenue which transfers knowledge from a source domain to a target domain without using any labels in the target domain. Contemporary techniques focus on extracting domain-invariant features using domain adversarial training. However, these techniques neglect to learn discriminative class boundaries in the latent representation space on a target domain and yield limited adaptation performance. To address this problem, we propose distance metric guided feature alignment (MetFA) to extract discriminative as well as domain-invariant features on both source and target domains. The proposed MetFA method explicitly and directly learns the latent representation without using domain adversarial training. Our model integrates class distribution alignment to transfer semantic knowledge from a source domain to a target domain. We evaluate the proposed method on fetal ultrasound datasets for cross-device image classification. Experimental results demonstrate that the proposed method outperforms the state-of-the-art and enables model generalization.

\end{abstract}
\section{Introduction}
Despite the success of deep neural networks (DNNs) for medical imaging applications~\cite{Dong2016,Chartsias2017,Liu2018,meng2019b,dou2020}, learning a task-specific model which generalizes to various medical datasets remains a challenge. This is due to the difference of feature distributions between different datasets, which is known as domain shift~\cite{joaquin2009}. In medical imaging, domain shift can result from different imaging modalities (\emph{e.g.}, magnetic resonance imaging and ultrasound) or different image acquisition devices. In this paper, we focus on model generalization between different image acquisition devices, transferring knowledge from a source device domain to a target device domain.

Fine-tuning DNNs on labelled data from the target domain is a possible solution but is often infeasible due to the need for sufficient manual annotations.
More importantly, fine-tuned models remain domain specific because performance gains do not propagate back to the source domain.
Deep domain adaptation has been widely studied for tackling the problem of domain shift by extracting domain-invariant features~\cite{Long2015,Ganin2016,Hausser2017}. Such approaches allow porting DNNs to the target domain without extensive annotation as well as preserving performance in both source and target domains. Unsupervised domain adaptation aims at transferring knowledge from a labeled source domain to an unlabeled target domain where both domains share a common label space~\cite{Ganin2016,Lee2019,meng2020}. This setting is important for real-world medical imaging scenarios, where data annotation is laborious, time consuming and requires rare expertise is available.

In this work, we propose distance metric guided feature alignment (MetFA) to learn a domain-invariant latent representations for model generalization in an unsupervised domain adaptation setting. We evaluate the proposed method on a challenging medical application, the classification of standardized diagnostic fetal ultrasound (US) view planes during prenatal screening. In many countries, fetal US is clinical routine for early detection of pathological development and informs subsequent decisions about treatment and delivery options~\cite{salomon2011}. However, domain shift caused by different acquisition devices 
and prohibitively expensive data annotation restricts the generalization of vanilla DNN classifiers. We show that MetFA enables unsupervised cross-device classification in fetal US. 

\noindent\textbf{Contribution.} The main contributions of this paper are: (1) We propose distance metric guided feature alignment (MetFA), which learns a shared latent representation space between a labeled source domain and an unlabeled target domain; 
(2) we develop a framework that jointly learns class distribution alignment and MetFA, which further transfers semantic knowledge from a source domain to a target domain for model generalization; 
(3) we utilize the proposed method for cross-device anatomical classification in fetal US, which is an important medical imaging application that inherently requires knowledge transfer between different device domains to facilitate the use of DNNs for large scale population screening.

\noindent\textbf{Related work.} \textit{\textbf{Unsupervised domain adaptation (UDA)}} mainly focuses on feature distribution alignment. Most UDA approaches explore an appropriate metric to measure the distance of feature distributions between two domains and subsequently train DNNs to minimize this distance~\cite{sohn2019}. Previous work such as Maximum Mean Discrepancy~\cite{Tzeng2014,Long2015} utilizes kernels to measure the discrepancy between representations. Recent research explores domain adversarial training, where a domain discriminator is used to estimate this discrepancy while a feature extractor tries to deceive the discriminator by learning domain-invariant representations~\cite{Bousmalis2016,Luo2017,Long2018}.
UDA has been applied to various medical imaging applications such as anatomical segmentation~\cite{Kamnitsas2016,Dong2018,Ouyang2019,Chen2019,Cai2019} and diagnostic classification~\cite{Huang2017,Bernard2018}. Most of these works utilize domain adversarial training for feature alignment. In contrast to these works, we explicitly manipulate the latent space to learn discriminative features. Our work is inspired by MiniMax Entropy (MME) proposed in~\cite{Saito2019}, which estimates domain-invariant prototypes and clusters target domain features around these prototypes in a \textit{semi-supervised domain adaptation setting}. In contrast to~\cite{Saito2019}, our method (1) embeds extracted features into a shared latent space with a fixed prior distribution before prototypes are estimated, 
and (2) simultaneously reduces intra-class variance while increasing inter-class variance across domains via cross-domain metric learning.

\textit{\textbf{Metric learning}} aims at learning embedded representations that cluster similar samples while separating dissimilar samples in latent space~\cite{Xing2003}. Previous metric learning methods measure feature similarity by learning a linear Mahalanobis distance~\cite{weinberger2009,Kostinger2012}. More recent works focus on deep metric learning, which learns non-linear relationships of data using DNNs with different losses, such as contrastive loss~\cite{chopra2005,Hadsell2006}, triplet loss~\cite{weinberger2009,chechik2010} and N-pair loss~\cite{Sohn2016}. Deep metric learning has shown great benefits for domain adaptation. For example, Sohn et al.~\cite{sohn2019} proposed a deep metric learning method for unsupervised domain adaptation in disjoint label space. Dou et al.~\cite{Dou2019} introduced deep metric learning for domain generalization. Most existing metric-learning-based domain adaptation methods only utilize metric learning on the labeled source domain and neglect the relationship between intra-class samples. In contrast to these methods, we introduce cross-domain metric learning to (1) jointly measure the similarity between samples in a labeled source domain and an unlabeled target domain and (2) learn metrics between different groups of intra-class samples.

\section{Method}
\begin{figure*}[tb]
 \centering
 \includegraphics[width=\textwidth, trim=8.7cm 9.6cm 9.3cm 3.9cm, clip]{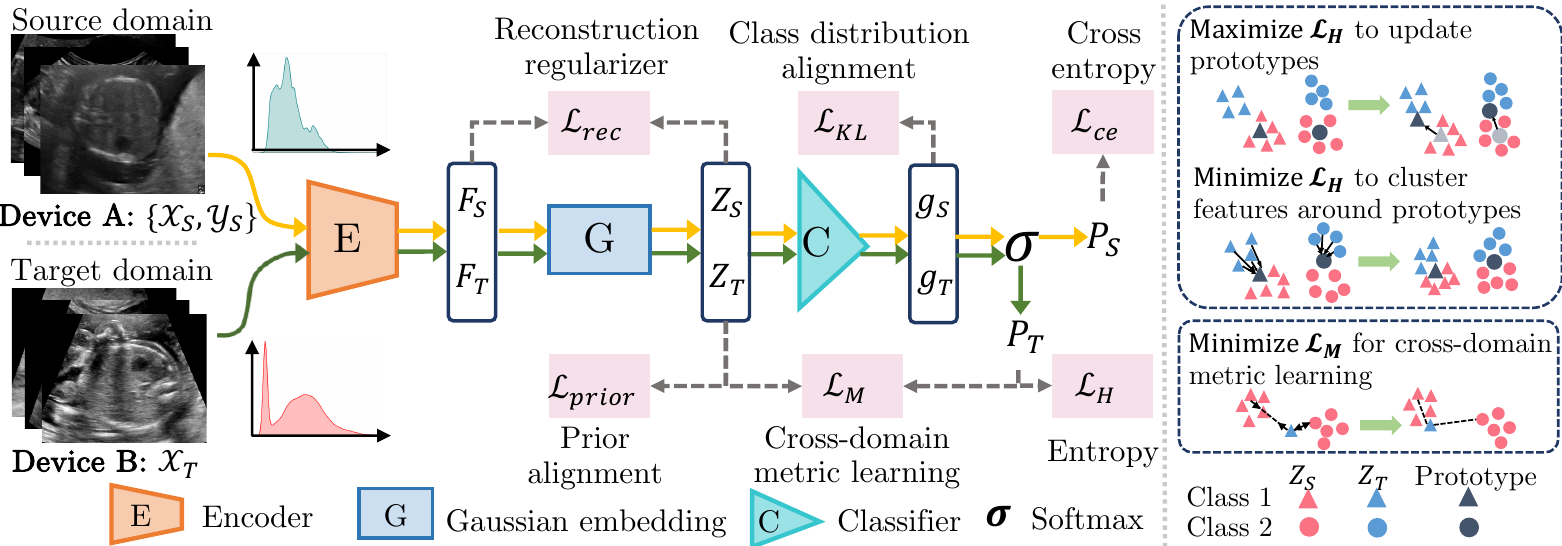}
 \caption{Left: An overview of the proposed method. Our method consists of (1) supervised classification on the labeled source domain (optimize $\mathcal{L}_{ce}$), (2) distance metric guided feature alignment (MetFA), which aligns features between both domains (optimize $\mathcal{L}_{prior}$, $\mathcal{L}_{H}$, $\mathcal{L}_{M}$, $\mathcal{L}_{rec}$), and (3) class distribution alignment, which preserves class relationships in both domains (optimize $\mathcal{L}_{KL}$). Right: Schematic illustration of $\mathcal{L}_H$ and $\mathcal{L}_{M}$ optimization.}
 \label{methed_outline}
\end{figure*}

We are given images and the corresponding labels from a source domain $\mathcal{D}_S=\{\mathcal{X}_S, \mathcal{Y}_S\}$ as well as unlabeled images from a target domain $\mathcal{D}_T=\{\mathcal{X}_T\}$. Both domains share a common label space and contain $M$ classes. Our goal is to classify unlabeled target domain data by aligning latent features of both domains. The proposed method contains three main parts (see Fig.~\ref{methed_outline}): (1) supervised classification on the labeled source domain, (2) distance metric guided feature alignment (MetFA) to transfer knowledge from the source domain to the target domain, and (3) class distribution alignment to preserve source domain class relationships in the target domain.  

\noindent\textbf{Classification.}
Classification in the unlabeled target domain is guided by the labeled source domain by sharing whole networks including an encoder $E$, a Gaussian embedding $G$ and a classifier $C$. The cross-entropy loss is 
\begin{equation}\label{Loss_ce}
\mathcal{L}_{ce}=-\mathbb{E}_{\{\mathbf{x},y\}\thicksim \{\mathcal{X}_S, \mathcal{Y}_S\}}\sum_{t=1}^{M}\mathbbm{1}[y=t]log(C(G(E(\mathbf{x})))).
\end{equation}
Classifier $C$ simultaneously predicts class distributions for the target domain as $P_T(\hat{y}|\mathbf{x})|_{\mathbf{x} \in \mathcal{X}_T}$ (abbreviated as $P_T$). This prediction will be utilized in MetFA.

\noindent\textbf{MetFA: distance metric guided feature alignment.}
Feature embedding is used to constrain features from both domains to lie in a shared latent space. 
In this latent space, class representations (prototypes) are estimated to extract domain-invariant features in each class, while cross-domain metric learning is introduced to further separate clusters of different classes in both domains.

\textit{\textbf{Feature embedding}} encourages features ($F_S$, $F_T$) extracted by an encoder $E$ to share the same fixed prior distribution in a latent space $\mathcal{Z}$, which is similar to distribution matching in a variational autoencoder~\cite{Kingma2014}. In our method, a Gaussian embedding $G$ is built to model $F_S$ and $F_T$ by a standard Gaussian distribution $\mathcal{N}(0,I)$. Specifically, $Z_i\sim q(\mathcal{Z}|\mathcal{X}_i)|i\in\{S,T\}$ is sampled from $\mathcal{N}(\mu_i, \Sigma_i)|i\in\{S,T\}$ with the reparameterization trick~\cite{Kingma2014}, where $\{\mu_i, \Sigma_i\}=G(F_i)|i\in\{S,T\}$ is implemented by a fully-connected layer. The prior alignment loss is the Kullback-Leibler (KL) divergence between $\mathcal{N}(0,I)$ and $\mathcal{N}(\mu_i, \Sigma_i)|i\in\{S,T\}$, which is 
\begin{equation}\label{Loss_prior}
\mathcal{L}_{prior}=D_{KL}(\mathcal{N}(\mu_S, \Sigma_S)\parallel\mathcal{N}(0,I))+D_{KL}(\mathcal{N}(\mu_T, \Sigma_T))\parallel\mathcal{N}(0,I)).
\end{equation}
In order to guarantee that embedded features are representative of the extracted features, we add a feature reconstruction loss $\mathcal{L}_{rec}$ as a regularizer:
\begin{equation}\label{Loss_rec}
\mathcal{L}_{rec}=\| F_S-\hat{Z}_S \|_2^2 + \| F_T-\hat{Z}_T \|_2^2,
\end{equation}
where $\hat{Z}_S$ and $\hat{Z}_T$ are respectively generated from $Z_S$ and $Z_T$ via a fully-connected layer and are the same dimension with $F_S$ and $F_T$. Feature embedding constrains distribution matching. In the absence of target domain labels, it is essential for subsequent feature alignment. 
However, feature embedding itself is unlikely to ensure that features are domain-invariant and discriminative between different classes. The rest of MetFA tackles this problem.

\textit{\textbf{Domain-invariant feature extraction}}
is motivated by Minimax Entropy (MME), proposed by Saito et al.~\cite{Saito2019}. Using unlabeled data in the target domain, MME learns a single domain-invariant prototype (a representation point) for each class in both domains and clusters target domain samples around these prototypes (see Fig.~\ref{methed_outline} upper right). We implement prototypes as the weights $\mathbf{W}$ of the last dense layer in the classifier $C$. 

Training MME contains two iterative steps. The first step is to move prototypes from source domain to target domain, which is maximizing the similarity between $\mathbf{W}$ and its input features ($H_T$). This similarity maximization is equivalent to maximizing the entropy of $\mathcal{X}_T$ with respect to $\mathbf{W}$, using 
\begin{equation}\label{Loss_H}
\mathcal{L}_{H}=-\mathbb{E}_{\mathbf{x}\sim \mathcal{X}_T}\sum_{i=1}^{M}p_T(\hat{y}=i|\mathbf{x})\log p_T(\hat{y}=i|\mathbf{x}), \:\: p_T\in P_T=\sigma(\frac{1}{\tau_0}\frac{\mathbf{W}^T H_T}{\|H_T\|}), 
\end{equation}
where $\sigma$ is a softmax function and $\tau_0$ is a temperature parameter.
The second step is to assign target domain features to the domain-invariant prototypes. To achieve this, $\mathcal{L}_H$ is minimized with respect to $E$, $G$ and $C\setminus\mathbf{W}$ ($C$ without $\mathbf{W}$). 

\textit{\textbf{Cross-domain metric learning}} is proposed to combine all samples in both domains and clusters samples that belongs to the same class and simultaneously separating samples from different classes. We define latent features of $\mathcal{X}_S$ and $\mathcal{X}_T$ (which are $Z_S$ and $Z_T$) respectively as support samples and query samples. In contrast to other metric learning loss where support and query sample are from the same domain with ground truth labels, in cross-domain metric learning, query samples are from the target domain with predicted labels and support samples are from the source domain with ground truth labels. The distance between query and support samples is minimized when they are from the same class and simultaneously maximized when they are from different classes (see Fig.~\ref{methed_outline} lower right). The metric loss is
\begin{equation}\label{Loss_M}
\mathcal{L}_{M}=\frac{1}{N}\sum_{i=1}^{M}\sum_{j=1}^{c_i^T}\log(1+\sum_{k\neq i}^{\mathclap{k\in[1,M]}}e^{d_j^i-d_j^k})=-\frac{1}{N}\sum_{i=1}^{M}\sum_{j=1}^{c_i^T}\log\frac{e^{d_j^i}}{e^{d_j^i}+\sum_{k\neq i}^{k\in[1,M]}e^{d_j^k}},
\end{equation}
where $N$ and $c_i^T$ are the number of all query samples and query samples from class $i$. Note that the labels of query samples are $P_T$ in Eq.~\ref{Loss_H}.
$d_j^i$ is the distance between a query sample $q_j^i$ and a same class support sample $s_t^i$. $d_j^k$ is the distance between $q_j^i$ and $s_t^k$ from different classes.
Considering the relationship between intra-class samples and using a hard mining strategy~\cite{ChenG2019}, we define $d_j^i$ and  $d_j^k$ as
\begin{equation}\label{loss_M_d}
\begin{split}
    &d_j^i=\max_t d(q_j^i,s_t^i), \; t \in [1, c_i^S], \: q_j^i \sim Z_T, \: s_t^i \sim Z_S, \\
    &d_j^k=\min_t d(q_j^i,s_t^k), \; t \in [1, c_k^S], \: q_j^i \sim Z_T, \: s_t^k \sim Z_S,
\end{split}
\end{equation}
where $c_i^S$ and $c_k^S$ are the number of support samples from class $i$ and class $k$. We use the squared Euclidean distance for $d(\cdot,\cdot)$ in Eq.~\ref{loss_M_d}.

\noindent\textbf{\textbf{Class distribution alignment.}} Apart from structuring a feature space for better class predictions, we want to further transfer semantic knowledge which is preserving class relationships between domains. 
Class distribution alignment is used for class relationship preservation between multiple labeled source domains in a domain generalization task~\cite{Dou2019}. In our method, we align class distributions between a labeled source domain and an unlabeled target domain.
We utilize the symmetrized KL-divergence to define the class distribution alignment loss
\begin{equation}\label{loss_kl}
\begin{gathered}
\mathcal{L}_{KL} = \frac{1}{M} \sum_{i=1}^M \Lambda [D_{KL}(\Bar{p}_i^S \parallel \Bar{p}_i^T) + D_{KL}(\Bar{p}_i^T \parallel \Bar{p}_i^S)],\\
\Bar{p}_i^S=\sigma(\frac{1}{\tau_1}\frac{1}{c_i^S}\sum_{y=i} g_{\mathbf{x}}^S)|_{(\mathbf{x},y) \sim \{\mathcal{X}_S, \mathcal{Y}_S\}}, \; \Bar{p}_i^T=\sigma(\frac{1}{\tau_1}\frac{1}{c_i^T}\sum_{\hat{y}=i} g_{\mathbf{x}}^T)|_{(\mathbf{x},\hat{y}) \sim \{\mathcal{X}_T, P_T(\mathbf{x})\}}.
\end{gathered}
\end{equation}
Here, $\Lambda=[c_1^T, c_2^T,...,c_M^T]$ contains the number of target domain samples predicted for each class. $\Bar{p}_i^S$ and $\Bar{p}_i^T$ are the $i^{th}$ class distributions in source and target domain.
$g_{\mathbf{x}}^S$ and $g_{\mathbf{x}}^T$ are the pre-softmax activations from classifier $C$ and $\tau_1$ is a temperature parameter.

\noindent\textbf{Optimization.}
The overall objective function of the proposed method is: 
\begin{equation}\label{Loss}
\begin{gathered}
    \min_{E,G,C\setminus\mathbf{W}}\{\mathcal{L}+\lambda_6\mathcal{L}_H\}, \quad \min_{\mathbf{W}}\{\mathcal{L}-\lambda_6\mathcal{L}_H\}, \\
    \text{with} \quad \mathcal{L}=\lambda_1\mathcal{L}_{ce}+\lambda_2\mathcal{L}_{prior}+\lambda_3\mathcal{L}_{M}+\lambda_4\mathcal{L}_{rec}+\lambda_5\mathcal{L}_{KL}.
\end{gathered}
\end{equation}
Here $\lambda_1$ to $\lambda_6$ are hyper-parameters chosen experimentally depending on the application. Our model is end-to-end trainable, with
$\mathbf{W}$ and the rest of the networks are trained in an alternating fashion according to Eq.~\ref{Loss}.
We apply L2 regularization ($\text{scale}=10^{-5}$) to all weights during training to prevent over-fitting and apply random image flipping as data augmentation. Our model is trained on a Nvidia Titan X GPU.

\section{Evaluation and results}

We evaluate the proposed method on 2D fetal US images acquired during routine prenatal screening. This US data is obtained by different imaging devices: Device A (GE Voluson E8) acquires $\sim12k$ images and device B (Philips EPIQ V7 G) acquires unpaired $\sim5.5k$ images. In both datasets, six anatomical standard planes have been selected by expert sonographers, including Four Chamber View (4CH), Abdominal, Left Ventricular Outflow Tract (LVOT), Right Ventricular Outflow Tract (RVOT), Femur and Lips. We evaluate our method in two scenarios where device A is the source domain while device B is the target domain, and vice versa. During training, the source domain is fully labeled and the target domain is unlabeled. In both scenarios, classes in the source domain are balanced for training and hyper-parameters $\lambda_1$ to $\lambda_6$ in Eq.~\ref{Loss} are $\lambda_1=10, \lambda_2=10^{-2}, \lambda_3=10^{-1}, \lambda_4=1, \lambda_5=10, \lambda_6=5$. $\tau_0$ in Eq.~\ref{Loss_H} is $0.05$ (same to~\cite{Saito2019}) and $\tau_1$ in Eq.~\ref{loss_kl} is $2$ (same to~\cite{Dou2019}). In the mini-batch during training, each class in the source domain contain 5 images and they are all used as support samples, $C_i^S=C_i^T=5$ in Eq.~\ref{loss_M_d}. We use Stochastic Gradient Descent (SGD) with momentum optimizer to update our model. 

\noindent\textbf{Comparison methods.} We evaluate a VGG network~\cite{meng2020} which contains an encoder $E$ and a classifier $C$ from the proposed method as a baseline. This baseline is trained on data only from the source domain (\textit{Source only}) to demonstrate the existence of domain shift. We compare the proposed method with the state-of-the-art domain-adaptation algorithms, including domain-adversarial training of neural networks (DANN)~\cite{Ganin2016}, adversarial discriminative domain adaptation (ADDA)~\cite{Tzeng2017} and semi-supervised domain adaptation via minimax entropy (MME)~\cite{Saito2019}. Note that for fair comparison, we use the MME model in an unsupervised learning paradigm. Additionally, given target domain labels, we show fine-tuned and fully-supervised classification on the target domain as references. Fine-tuned classification is pre-trained on the labeled source domain and fine-tuned on the labeled target domain. This fine-tuned model is evaluated on both source and target domains. Fully-supervised classification is trained from scratch on the labeled target domain and evaluated on the target domain. 

\noindent\textbf{Ablation study.} We further explore the effectiveness of different components in the proposed method by removing different loss components: UDA-MetFA-I: only contains $\mathcal{L}_{ce}$, $\mathcal{L}_{prior}$ and $\mathcal{L}_{H}$; UDA-MetFA-II: UDA-MetFA-I plus $\mathcal{L}_{M}$; UDA-MetFA-III: UDA-MetFA-II plus $\mathcal{L}_{KL}$; UDA-MetFA-IV: UDA-MetFA-II plus $\mathcal{L}_{rec}$; UDA-MetFA-V: contains all components.

\noindent\textbf{Results.} Table~\ref{1to2_table} shows the experimental results of baselines and the ablation study where device A is the source domain and device B is the target domain. From this table, we observe that the UDA-MetFA-V model outperforms other baselines. In the target domain, UDA-MetFA-V achieves an average F1-score of $0.7713$ while the highest average F1-score of other baselines is $0.4398$ (MME~\cite{Saito2019}). UDA-MetFA-I greatly outperforms MME~\cite{Saito2019} in the target domain, demonstrating the importance of feature embedding in the proposed method. UDA-MetFA-V performs better than other ablation models in the target domain, illustrating the effectiveness of all compoents in the proposed method. Furthermore, the results of Fine-tuned and \textit{source only} in the source domain indicate that the fine-tuned model remains less generalizable, whereas the proposed method (UDA-MetFA-V) enables model generalization with improved classification performance in both source and target domains. 

\begin{figure*}[htb]
 \centering
 \setcounter{subfigure}{0}
 \subfloat[Confusion Matrix]{
 \begin{tabular}{@{\hspace{0\tabcolsep}}c@{\hspace{-1\tabcolsep}}c}
  \stackunder{\includegraphics[height=2.7cm, trim=0cm 0cm 2cm 0cm, clip]{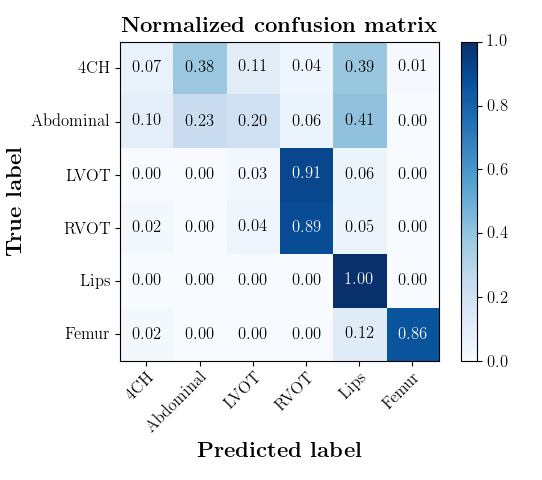}}{~\scalebox{1}{MME~\cite{Saito2019}}} &
  \stackunder{\includegraphics[height=2.7cm, trim=0cm 0cm 0cm 0cm, clip]{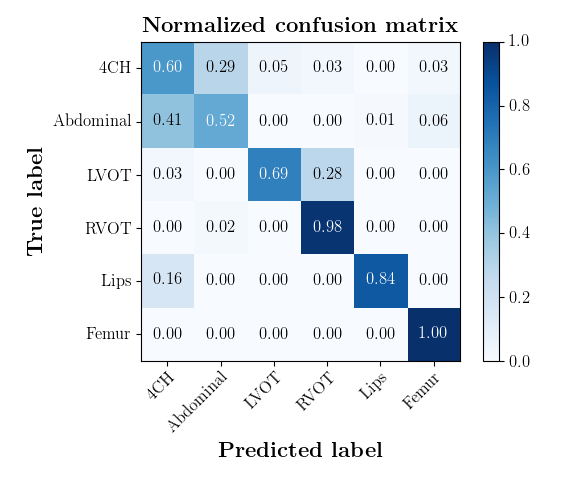}}{~\scalebox{1}{UDA-MetFA-V}} 
  \end{tabular}
  }
  \hfill
  \hspace{-0.5cm}
  \setcounter{subfigure}{1}
 \subfloat[t-SNE visualization]{
 \begin{tabular}{c@{\hspace{-1\tabcolsep}}c@{\hspace{0\tabcolsep}}}
  \stackunder{\includegraphics[height=2.6cm,trim=1.6cm 1.1cm 0cm 1.1cm, clip]{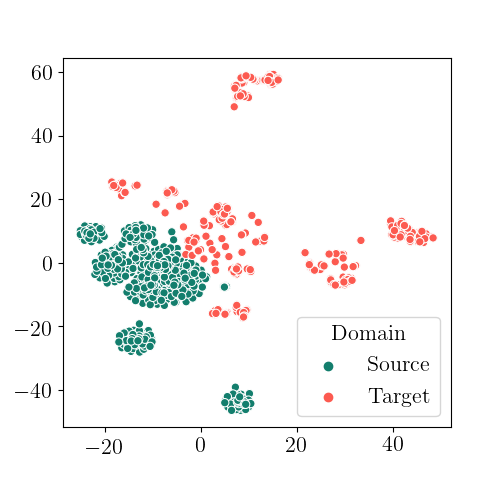}}{~\scalebox{1}{MME~\cite{Saito2019}}} &
  \stackunder{\includegraphics[height=2.6cm,trim=1.6cm 1.1cm 0cm 1.1cm, clip]{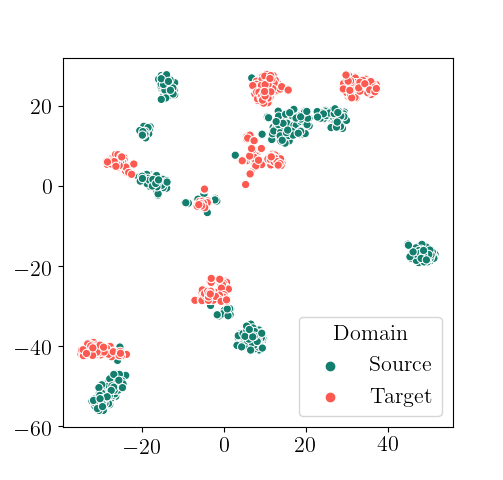}}{~\scalebox{1}{UDA-MetFA-V}}
  \end{tabular}
  }
  \caption{Comparison of MME~\cite{Saito2019} and UDA-MetFA-V on (a) confusion matrix of target domain (device B) and (b) t-SNE plot of extracted test data features. 
  }
  \label{CM_tsne}
\end{figure*}

We further compare MME (best baseline in Table~\ref{1to2_table}) with the proposed method (UDA-MetFA-V) in confusion matrices and t-SNE plots.
Fig.~\ref{CM_tsne}(a) demonstrates that our method extracts more discriminative features for better classification, especially on easily confused anatomies (\emph{e.g.}, LVOT vs. RVOT). 
Fig.~\ref{CM_tsne}(b) shows that for UDA-MetFA-V, target features $Z_T$ are closer to source features $Z_S$ while features of different classes are more separated. This indicates that the proposed MetFA benefits the extraction of discriminative and domain-invariant features. 

Table~\ref{2to1_table} shows the results of comparison methods and the proposed method (UDA-MetFA-V) on switched domains, where device B is the source domain and device A is the target domain. We observe that UDA-MetFA-V outperforms the state-of-the-art in both source and target domains, demonstrating that our method is capable of successfully transferring knowledge from source domain to target domain as well as improving model generalization. 

\begin{table}[tb]
\centering
\caption{Comparison of \textit{Source only}, the state-of-the-art and ablation study (UDA-MetFA- I to V) for fetal US anatomical classification with \textbf{device A as source domain and device B as target domain}. Fine-tuned and Fully-supervised are reference results given target domain labels. Best results in bold. }
\label{1to2_table}
\begin{tabular}{cccc|ccc}
\toprule[1.2pt]
\multirow{2}{*}{Methods}                            & 
\multicolumn{3}{c|}{\textbf{S: Device A}}           &
\multicolumn{3}{c}{\textbf{T: Device B}}           \\
\cmidrule{2-7}
~~~~~                                               &
F1-score                                                & 
Recall                                              & 
Precision                                           &
F1-score                                                & 
Recall                                              & 
Precision                                           \\
\midrule
Source only                                   &
0.8782                                              &
0.8800                                              &
0.8786                                              &
0.2455                                              &
0.3400                                              &
0.3070                                              \\
ADDA~\cite{Tzeng2017}                                          &
0.8841                                              &
0.8850                                             &
0.8860                                             &
0.1377                                              &
0.2050                                              &
0.1623                                               \\
DANN~\cite{Ganin2016}                                          &
0.8321                                              &
0.8350                                             &
0.8564                                             &
0.3390                                              &
0.3650                                              &
0.3756                                              \\
MME~\cite{Saito2019}                                          &
0.8876                                              &
0.8900                                              &
0.8914                                              &
0.4398                                              &
0.5133                                               &
0.4565                                              \\
\midrule
UDA-MetFA-I                                          &
0.8894                                              &
0.8900                                              &
0.8911                                              &
0.5255                                              &
0.5550                                               &
0.5599                                              \\
UDA-MetFA-II                                          &
0.8951                                              &
0.8967                                              &
0.8997                                              &
0.5959                                              &
0.6400                                               &
0.6359                                              \\
UDA-MetFA-III                                          &
\textbf{0.9202}                                     &
\textbf{0.9200}                                     &
\textbf{0.9207}                                     &
0.6301                                              &
0.6850                                               &
0.6143                                              \\
UDA-MetFA-IV                                          &
0.8970                                              &
0.8967                                              &
0.8986                                              &
0.6930                                              &
0.7067                                               &
0.7011                                              \\
UDA-MetFA-V                                  &
0.8990                                              &
0.9000                                             &
0.9027                                             &
\textbf{0.7713}                                     &
\textbf{0.7717}                                    &
\textbf{0.7874}                                     \\
\midrule
Fine-tuned                                  &
0.7987                                              &
0.8050                                             &
0.8140                                             &
0.7114                                              &
0.7150                                             &
0.7373                                             \\
Fully-supervised                                  &
--                                              &
--                                             &
--                                             &
0.5919                                              &
0.6100                                             &
0.6576                                             \\
\bottomrule[1.2pt]
\end{tabular}
\end{table}

\begin{table}[tb]
\centering
\caption{Comparison of baselines and UDA-MetFA-V with \textbf{device B as source domain and device A as target domain}. Best results in bold. 
}
\label{2to1_table}
\begin{tabular}{cccc|ccc}
\toprule[1.2pt]
\multirow{2}{*}{Methods}                            & 
\multicolumn{3}{c|}{\textbf{S: Device B}}           &
\multicolumn{3}{c}{\textbf{T: Device A}}           \\
\cmidrule{2-7}
~~~~~                                               &
F1-score                                                & 
Recall                                              & 
Precision                                           &
F1-score                                                & 
Recall                                              & 
Precision                                           \\
\midrule
Source only                                   &
0.5919                                              &
0.6100                                             &
0.6576                                              &
0.2854                                              &
0.3300                                              &
0.3555                                              \\
DANN~\cite{Ganin2016}                                          &
0.5198                                              &
0.5450                                             &
0.5451                                             &
0.3318                                              &
0.3500                                              &
0.3450                                              \\
MME~\cite{Saito2019}                                          &
0.3776                                              &
0.4183                                              &
0.4500                                              &
0.1520                                              &
0.1883                                               &
0.2101                                              \\
UDA-MetFA-V                                  &
\textbf{0.7101}                                     &
\textbf{0.7150}                                     &
\textbf{0.7441}                                     &
\textbf{0.5776}                                     &
\textbf{0.5550}                                    &
\textbf{0.6303}                                     \\
\midrule
Fully-supervised                                  &
--                                              &
--                                             &
--                                             &
0.8782                                              &
0.8800                                              &
0.8786                                             \\
\bottomrule[1.2pt]
\end{tabular}
\end{table}

\noindent\textbf{Discussion.} Domain adaptation is commonly used to transfer a performant, task-specific model from a source domain to a target domain. However, the DNNs learning ability in a source domain can limit this ability in a target domain. This may explain the lower classification performance of the proposed method compared with a fully-supervised method in the target domain in Table~\ref{2to1_table}, where the classification of the source domain is relatively low (see Source only). 
Current UDA methods rarely discuss the performance of DNNs in the source domain. 
From Table.~\ref{2to1_table}, we observe that tracking the source domain performance can be potentially used for data selection during model improvement in the source domain. 
A limitation of our method is the empirical hyper-parameters selection. For a specific application, we adjust hyper-parameters according to their importance and select the best combination with grid search. Meta-learning~\cite{feurer2015} will be explored in future work to allow automatic hyper-parameter selection.

\section{Conclusion}
In this paper, we discuss the problem of model generalization for unsupervised domain adaption. We propose metric learning for improved feature alignment (MetFA)\footnote{codes in https://github.com/qingjie99/MetFA} to extract discriminative and domain-invariant features across domains. MetFA explicitly structures latent representations without using domain adversarial training. Our model integrates class distribution alignment for transferring semantic knowledge from a source domain to a target domain. Experiments on cross-device fetal US screening images demonstrate the effectiveness and practical applicability of our method compared with the state-of-the-art.

\subsubsection*{Acknowledgments.} We thank the Wellcome Trust IEH Award [102431], Nvidia (GPU donations).
%
%
%
\bibliographystyle{splncs04}

\newpage

\appendix
\textbf{\LARGE{Appendices}}
\section{Examples of Ultrasound Images}

We show more examples of ultrasound images acquired from different image acquisition devices.

\begin{figure}
    \centering
 \subfloat[4CH]{
 \begin{tabular}{c@{\hspace{3\tabcolsep}}c}
  \raisebox{0.5\height}{\rotatebox[origin=c]{0}{\makecell{~\scalebox{0.8}{\textbf{Device A}}}}}  &
 \raisebox{0.5\height}{\rotatebox[origin=c]{0}{\makecell{~\scalebox{0.8}{\textbf{Device B}}}}}  \\
  \includegraphics[height=2.2cm]{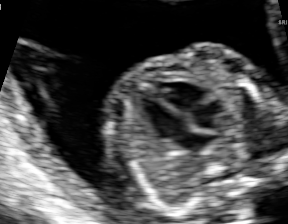} &
  \includegraphics[height=2.2cm]{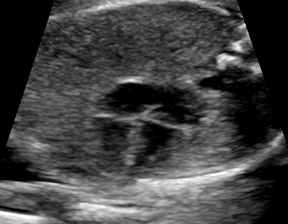} 
  \end{tabular}
  }
  \hfill
  \hspace{-0.5cm}
  \subfloat[Abdominal]{
 \begin{tabular}{c@{\hspace{3\tabcolsep}}c}
  \raisebox{0.5\height}{\rotatebox[origin=c]{0}{\makecell{~\scalebox{0.8}{\textbf{Device A}}}}}  &
 \raisebox{0.5\height}{\rotatebox[origin=c]{0}{\makecell{~\scalebox{0.8}{\textbf{Device B}}}}}  \\
  \includegraphics[height=2.2cm]{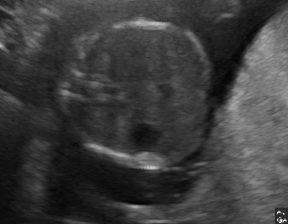} &
  \includegraphics[height=2.2cm]{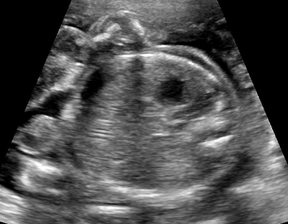} 
  \end{tabular}
  }
  \hfill
 \subfloat[Femur]{
 \begin{tabular}{c@{\hspace{3\tabcolsep}}c}
  \raisebox{0.5\height}{\rotatebox[origin=c]{0}{\makecell{~\scalebox{0.8}{\textbf{Device A}}}}}  &
 \raisebox{0.5\height}{\rotatebox[origin=c]{0}{\makecell{~\scalebox{0.8}{\textbf{Device B}}}}}  \\
  \includegraphics[height=2.2cm]{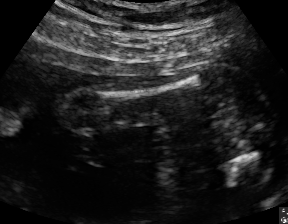} &
  \includegraphics[height=2.2cm]{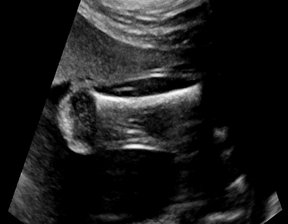} 
  \end{tabular}
  }
  \hfill
  \hspace{-0.5cm}
  \subfloat[Lips]{
 \begin{tabular}{c@{\hspace{3\tabcolsep}}c}
  \raisebox{0.5\height}{\rotatebox[origin=c]{0}{\makecell{~\scalebox{0.8}{\textbf{Device A}}}}}  &
 \raisebox{0.5\height}{\rotatebox[origin=c]{0}{\makecell{~\scalebox{0.8}{\textbf{Device B}}}}}  \\
  \includegraphics[height=2.2cm]{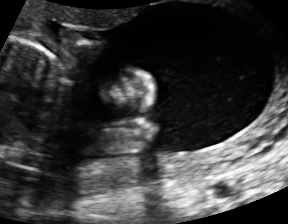} &
  \includegraphics[height=2.2cm]{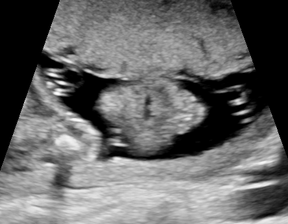} 
  \end{tabular}
  }
  \hfill
 \subfloat[LVOT]{
 \begin{tabular}{c@{\hspace{3\tabcolsep}}c}
  \raisebox{0.5\height}{\rotatebox[origin=c]{0}{\makecell{~\scalebox{0.8}{\textbf{Device A}}}}}  &
 \raisebox{0.5\height}{\rotatebox[origin=c]{0}{\makecell{~\scalebox{0.8}{\textbf{Device B}}}}}  \\
  \includegraphics[height=2.2cm]{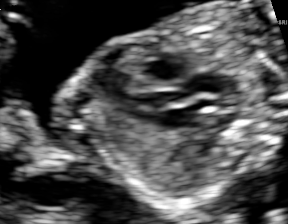} &
  \includegraphics[height=2.2cm]{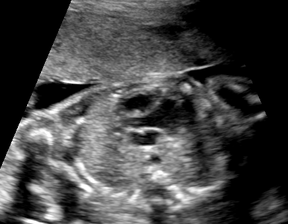} 
  \end{tabular}
  }
  \hfill
  \hspace{-0.5cm}
  \subfloat[RVOT]{
 \begin{tabular}{c@{\hspace{3\tabcolsep}}c}
  \raisebox{0.5\height}{\rotatebox[origin=c]{0}{\makecell{~\scalebox{0.8}{\textbf{Device A}}}}}  &
 \raisebox{0.5\height}{\rotatebox[origin=c]{0}{\makecell{~\scalebox{0.8}{\textbf{Device B}}}}}  \\
  \includegraphics[height=2.2cm]{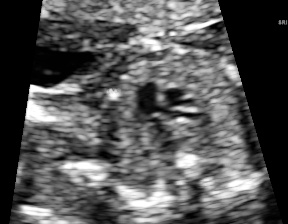} &
  \includegraphics[height=2.2cm]{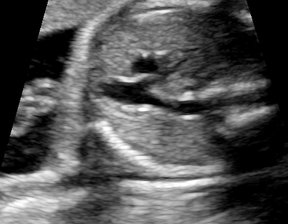} 
  \end{tabular}
  }
    \caption{Examples of ultrasound images acquired by different image acquisition devices. Device A is \textit{GE Voluson E8} and device B is \textit{Philips EPIQV7 G}.}
    \label{usex}
\end{figure}

\newpage

\section{Split of training data}

\begin{table}[h]
\centering
\caption{The number of images in each class for training. In the first scenario (S: device A, T: device B), images in device A are used as labeled data and images in device B are unlabeled. In the second scenario (S: device B, T: device A), images in device B are labeled and images in device A are unlabeled.
}
\label{trcls}
\begin{tabular}{c|cccccc}
    \toprule
    ~~~  &
    4CH  &
    Abdominal  &
    Femur  &
    Lips  &
    LVOT  &
    RVOT  \\
    \midrule
    Device A  &
    700  &
    700  &
    700  &
    700  &
    700  &
    700  \\
    Device B  &
    828  &
    728  &
    815  &
    600  &
    328  &
    559  \\
    \bottomrule
    \end{tabular}
\end{table}

\end{document}